\documentclass[sigconf]{acmart}
\usepackage{subcaption}
\usepackage{multirow}
\usepackage{tabularx}
\usepackage{booktabs}
\usepackage{color}
\usepackage{balance}

\newcommand{\eg}{\emph{e.g.}}
\newcommand{\ie}{\emph{i.e.}}

\newcolumntype{C}{>{\centering\arraybackslash}X}
\AtBeginDocument{%
  \providecommand\BibTeX{{%
    \normalfont B\kern-0.5em{\scshape i\kern-0.25em b}\kern-0.8em\TeX}}}

\copyrightyear{2021}
\acmYear{2021}
\setcopyright{acmcopyright}\acmConference[MM '21]{Proceedings of the 29th ACM
International Conference on Multimedia}{October 20--24, 2021}{Virtual Event, China}
\acmBooktitle{Proceedings of the 29th ACM International Conference on Multimedia
(MM '21), October 20--24, 2021, Virtual Event, China}
\acmPrice{15.00}
\acmDOI{10.1145/3474085.3475503}
\acmISBN{978-1-4503-8651-7/21/10}

\begin{document}
\fancyhead{}

\title{Identity-aware Graph Memory Network for Action Detection}


\author{Jingcheng Ni}
\affiliation{%
  \institution{State Key Laboratory of Software Development Environment, SCSE, Beihang University}
  \city{Beijing}
  \country{China}
  \postcode{100191}
}
\email{kiranjc@buaa.edu.cn}

\author{Jie Qin}
\affiliation{%
  \institution{College of Computer Science and Technology, Nanjing University of Aeronautics and Astronautics}
  \city{Nanjing}
  \country{China}
}
\email{qinjiebuaa@gmail.com}

\author{Di Huang}
\authornote{indicates the corresponding author.}
\affiliation{%
  \institution{State Key Laboratory of Software Development Environment, SCSE, Beihang University}
  \city{Beijing}
  \country{China}
  \postcode{100191}
}
\email{dhuang@buaa.edu.cn}

\renewcommand{\shortauthors}{Ni, et al.}

\begin{abstract}
  Action detection plays an important role in high-level video understanding and media interpretation. Many existing studies fulfill this spatio-temporal localization by modeling the context, capturing the relationship of actors, objects, and scenes conveyed in the video. However, they often universally treat all the actors without considering the consistency and distinctness between individuals, leaving much room for improvement. In this paper, we explicitly highlight the identity information of the actors in terms of both long-term and short-term context through a graph memory network, namely identity-aware graph memory network (IGMN). Specifically, we propose the hierarchical graph neural network (HGNN) to comprehensively conduct long-term relation modeling within the same identity as well as between different ones. Regarding short-term context, we develop a dual attention module (DAM) to generate identity-aware constraint to reduce the influence of interference by the actors of different identities. 
  Extensive experiments on the challenging AVA dataset demonstrate the effectiveness of our method, which achieves state-of-the-art results on AVA v2.1 and v2.2. 
\end{abstract}




\maketitle

\begin{figure}[ht]
   \centering
   \includegraphics[width=\linewidth]{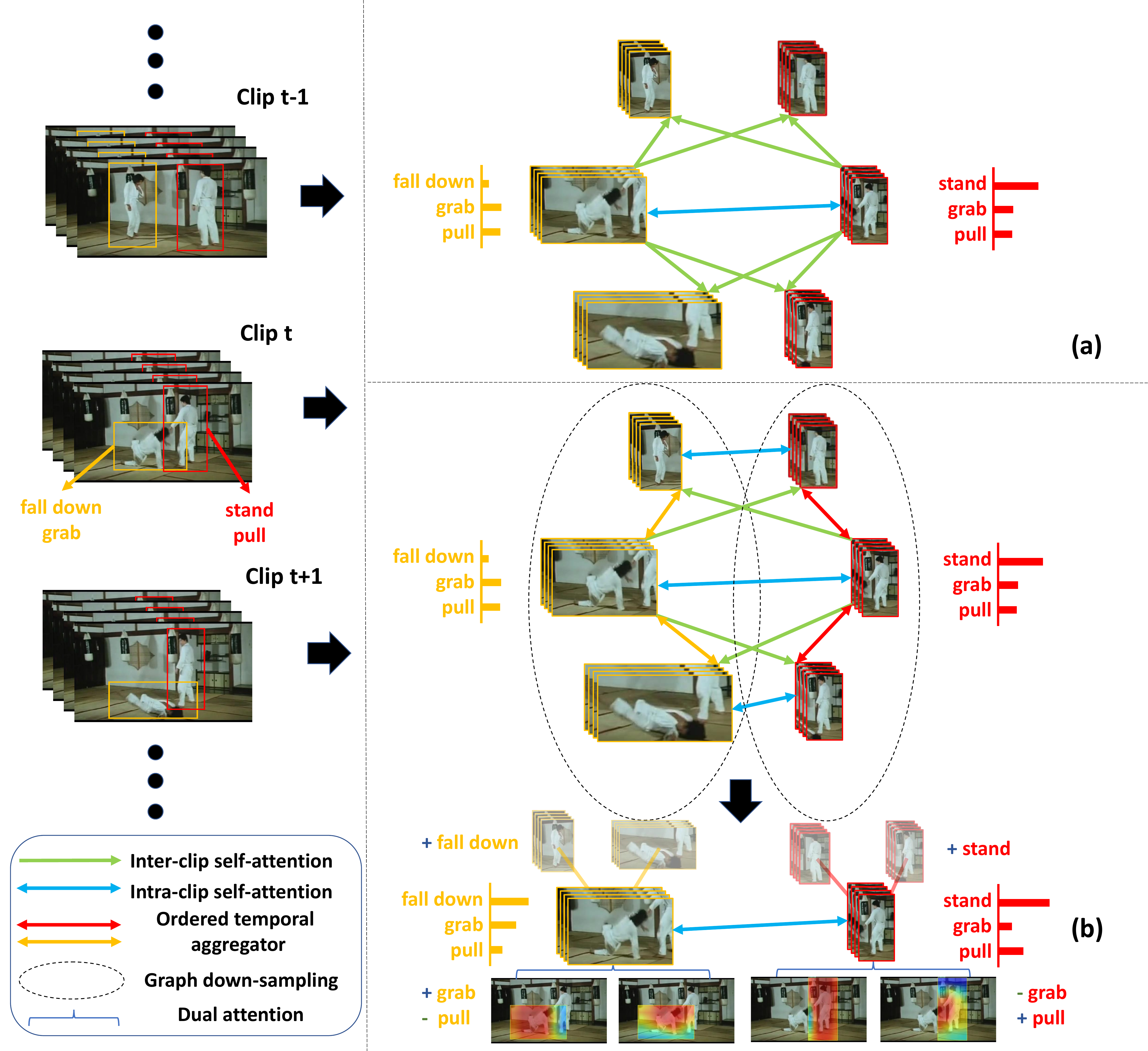}
   \caption{Intuition of IGMN. The goal is to localize and recognize the actions at timestamp $t$ based on short-term (within clip $t$) and long-term (across different clips) context. (a) Previous methods model relations of actors by self-attention without considering their identities. (b) We explicitly highlight the identity cues. For long-term context, ordered temporal aggregator and graph down-sampling are applied to extract multi-scale identity-aware interactions. For short-term context, we develop a dual attention module to locate identity-aware discriminative regions.}
   \label{fig:1}
\end{figure}

\section{Introduction}

Action detection (also known as spatio-temporal action localization) aims to locate actors and categorize actions within a given video. It is an essential step toward high-level video understanding and media interpretation. Existing methods mainly follow the dense sampling strategy, and features are built from bounding boxes, where feature maps of target actors are cropped from the whole video both spatially and temporally. The release of the latest benchmark \cite{gu2018ava} highlights the challenge of action detection in more complex scenarios. In this case, the cues of individual actors encoded by local features in current solutions are not adequate, and those of surrounding context are also necessary. As Fig. 1 (a) depicts, to accurately recognize the action of the person in yellow bounding boxes, it is desired to include the interaction with the person marked by red bounding boxes (\emph{w.r.t.} ``grab'') and the ordered temporal dynamics in the tracklet (\emph{w.r.t.} ``fall down'').  

In the literature, many studies \cite{sun2018actor, wu2019long, tang2020asynchronous, pan2020actor, wu2020context} have discussed context modeling for action detection. \cite{sun2018actor} shows preliminary attempts to capture the relationship between actors and scenes using specifically designed convolutional layers or certain attention mechanisms. However, this phase only works within video clips (denoted as \textbf{short-term context}), making it problematic to actions of long sequences. \cite{wu2019long} delivers a substantial extension to capture context among video clips (denoted as \textbf{long-term context}), which presents a self-attention based memory network to analyze the relationship of different actors. The follow-up alternatives enhance short-term (intra-clip) and/or long-term (inter-clip) context modeling for performance improvements. For instance, \cite{wu2020context, pan2020actor} ameliorate the former by enlarging spatial receptive fields and building high-order descriptions of actors and scenes, while \cite{li2019long, tang2020asynchronous} strengthen the latter through considering more detailed relationships in the network structures of graph convolutions and dense connections. However, the methods mentioned above commonly suffer from an intrinsic issue that the identity information of actors is neglected, which incurs actor confusion in stably learning both short-term and long-term context.

Specifically, long-term context refers to relations across clips. Such relations are mainly divided into intra-actor and inter-actor ones, where identity consistency is rather crucial. Taking the actor in yellow bounding boxes in Fig. 1 (b) as an example: intra-actor relation (linked by yellow edges) offers temporal clues to recognize the action of ``fall down'' and inter-actor relation (linked by green edges) captures interaction to recognize the action of ``grab''. Fig. 1 (a) shows the widely adopted pipeline \cite{wu2019long} and it is observed that the single attention-based strategy fails to model both inter-actor and intra-actor relations in the presence of temporal changes of actors. This is evidenced by the comparable scores in \cite{wu2020context} achieved by the attention-based aggregator and the simple average pooling. 
Short-term context focuses on relations within a clip and the identity issue also appears.
As other persons may exist in the proposals of the target actor, particularly in actions with human-human interactions, \emph{e.g.}, pushing and hitting, this interference tends to induce wrong positions for feature learning. As in Fig. 1 (b), the actor in red boxes is with the action of ``pull'', but the prediction suggests ``grab'' due to region overlapping with the actor in yellow boxes.

To solve the identity confusion problem, this paper proposes a novel approach to video action detection, namely identity-aware graph memory network (IGMN). By taking the actor features saved in the memory bank as nodes, the identity-aware graph is first built according to trajectories and timestamps. Two side branches are then designed for long-term context and short-term context learning, respectively. The long-term branch is the hierarchical graph neural network (HGNN). Unlike previous counterparts, inter-actor and intra-actor relations are separately processed in each hierarchical layer, and relation modeling is thus more comprehensively conducted. Another key design for HGNN is the down-sampling operator, highlighting the connections between different layers and conveying multi-scale temporal relations. The short-term branch is a dual attention module (DAM), whose purpose is to generate identity-aware constraint to reduce the interference by other actors and produce discriminative local context representation. To this end, we present the semantic attention and identity attention to replace the original one, corresponding to semantic clues and identity mask. The actor features are further weighted by both the attentions for context learning. The outputs of the two branches are finally merged for prediction. 

In summary, our key contributions include:

\textbf{(1)} We propose a graph memory network for action detection, which explicitly highlights the identity information in both long-term and short-term context modeling. 

\textbf{(2)} We develop HGNN and DAM to model multi-scale identity-aware relations in long-term context and emphasize identity-aware discriminative regions in short-term context. 


\textbf{(3)} Extensive experiments on the challenging AVA dataset
demonstrate the superiority of IGMN and verify the necessity of incorporating identity cues into action detection.

\begin{figure*}[t]
  \centering
  \begin{subfigure}{\linewidth}
    \centering
    \includegraphics[width=\linewidth]{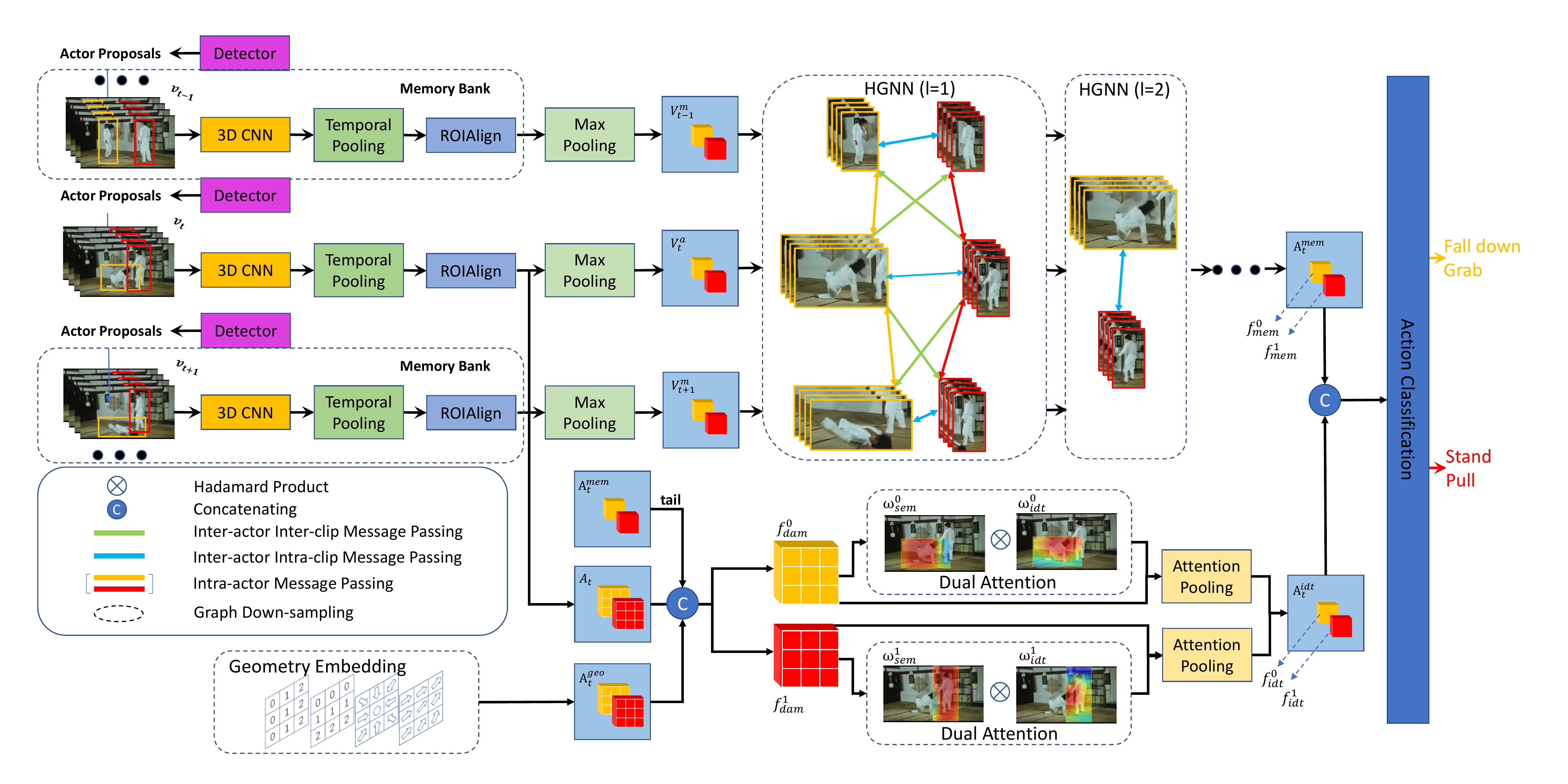}
  \end{subfigure}
  \caption{Overview of the proposed framework. Given an input video clip at timestamp $t$, we take actor features within a temporal range from the memory bank (here we choose three consecutive clips for simplicity). HGNN models long-term context by three message passing operations: intra-actor (yellow/red edges), inter-actor intra-clip (blue edges), and inter-actor inter-clip (green edges) message passing. We execute these operations in a single HGNN layer and introduce graph down-sampling between different layers. As for short-term context modeling, we apply DAM to learn two distinct attention maps.
  The updated actor features from the two modules are finally merged for action classification.
  }
  \label{fig:2}
\end{figure*}

\section{Related Work}
\textbf{Video Classification.}
Compared to image classification that only relies on static features, video classification requires spatio-temporal ones from consecutive frames and is thus more challenging. 
Various 3D CNNs \cite{ji20123d, tran2015learning, feichtenhofer2020x3d} have been developed in the hope that the models can learn hierarchical motion patterns as in the image domain. Due to time-consuming computation in 3D convolutions, some works \cite{xie2018rethinking, lin2019tsm} attempt to decompose 3D kernels into separate 2D spatial and 1D temporal ones for smaller size and better optimization. In the meantime, the sampling strategy is also crucial to this task. For example, dense sampling captures accurate short-term dynamics and sparse sampling \cite{wang2018temporal} extracts long-term dependencies. Several recent studies also investigate adaptive sampling for higher efficiency \cite{korbar2019scsampler, fayyaz20203d}. 

\noindent\textbf{Spatio-Temporal Action Detection.}
Action detection aims at not only recognizing action labels but also localizing actors in time and space, which is more difficult than action classification. There are two major research lines in supervised action detection, supported by different benchmarks. The first line concentrates on localization accuracy, measured by the video mAP on UCF101-24 \cite{soomro2012ucf101} and J-HMDB-21 \cite{jhuang2013towards}. \cite{kalogeiton2017action, li2020actions, song2019tacnet} adapt different object detection paradigms to action detection. To generate the whole action tube from clip-level proposals, a number of linking algorithms are proposed for post-processing. \cite{li2020finding} introduces a top-down method without the connecting procedure, taking the entire video as input to localize the temporal window of an action directly. The second line focuses on classification accuracy, driven by AVA \cite{gu2018ava} with complex scenes.  
Most current solutions depend on context details and relations on this challenging dataset.
To model short-term context within one video clip, \cite{ulutan2020actor} employs actor-conditioned attention maps to capture relations between actors and scenes. ACAR-Net \cite{pan2020actor} further extends this relation into a high-order form. Context-Aware RCNN \cite{wu2020context} exploits the RCNN framework \cite{girshick2014rich} for scale invariance.
For long-term context across video clips, LFB \cite{li2019long} firstly uses a network to save actor features in the memory bank and retrieves them by self-attention \cite{wang2018non}. AIA \cite{tang2020asynchronous} proposes a dense-serial structure to model complex interactions. 
Though the methods above take spatio-temporal contextual information into account, they overlook the identity issue in modeling relations. In this paper, we demonstrate how identity cues lead to better action detection performance.

\section{Proposed Method}
Action detection aims to localize persons both spatially and temporally in a video and meanwhile recognize their actions. As previously mentioned, sufficient use of context is of vital importance to decent results. In this paper, we propose identity-aware graph memory network (IGMN) to model both long- and short-term context. Unlike existing works that treat all actors similarly (\eg, using self-attention), IGMN highlights the identity information for improved detection performance. As shown in Fig. \ref{fig:2}, IGMN consists of two branches for context modeling, \ie, hierarchical graph neural network (HGNN) and dual attention module (DAM). In the following, we first introduce how to build clip-level actor features from the backbone network and the memory bank. Subsequently, HGNN and DAM are presented to capture long- and short-term identity-aware context, respectively. Finally, a joint loss function is introduced for model training.

\subsection{Actor Feature Extraction}
We follow the RoI-Pooling and dense-sampling based pipeline \cite{tang2020asynchronous} to extract actor features. A long video is split into ordered short clips $[v_1, v_2, ..., v_T]$, and we take a video clip ${v_t}$ as input and detect the actions of its center frame at timestamp ${t}$. The whole video can be processed by successively handling all the clips with their center frames as the key frames. To perform localization on the center frame of the clip at timestamp ${t}$, we apply the off-the-shelf Faster-RCNN detector \cite{ren2015faster} to generate $N_t$ actor proposals with 2D bounding box coordinates $b_t^i = (x^i_t, y^i_t, h^i_t, w^i_t)$ for the $i$-th actor in $v_t$, where $(x^i_t, y^i_t)$ is the top-left coordinate of the box and $(h^i_t, w^i_t)$ is the height and width. To better represent temporal data, a popular video backbone (\ie, SlowFast \cite{feichtenhofer2019slowfast})
is used to compute the 3D feature volume of the input clip. Then, we perform average pooling along the temporal dimension, resulting in a feature map ${I \in \mathbb{R}^{C \times H \times W}}$, where $C$ indicates the number of channels, and $H$ and $W$ denote the height and width, respectively. This 2D feature map is further processed by ROIAlign \cite{he2017mask} to produce the actor features $A_t = [f_t^1,...,f_t^i,...,f_t^{N_t}]$ in the video clip ${v_t}$, where $f_t^i \in \mathbb{R}^{C \times K \times K}$ represents the cropped feature of the $i$-th actor and $K$ is the feature map size. To model long-term temporal context across different clips, we employ the memory bank that keeps track of the extracted actor features in the whole video $\{A_t\}_{t=1}^{T}$ as in \cite{wu2019long}. When processing ${v_t}$, the corresponding consecutive memory features $M_t = [A_{t-L},...,A_{t-1},A_{t+1},...,A_{t+L}]$ are taken from the memory bank, where $L$ controls the size of the temporal window. Please refer to \cite{wu2019long} for more details.


\subsection{Hierarchical Graph Neural Network}
The proposed HGNN explicitly captures actor relationships in terms of the long-term context. Specifically, HGNN adopts a multi-layer structure, in which each layer models the relationships based on three different kinds of message passing. Inspired by the previous hierarchical structure \cite{cai2018cascade}, we enhance the distinctiveness of individual layers by introducing down-sampling on the graph,
enabling the interactions between different temporal scales.

\subsubsection{\textbf{Graph Construction.}}
Given a video clip ${v_t}$, we define the actor node set as $V_t^a = \{ {\nu}_{t,1}^a,\ ...,{\nu}_{t,i}^a,\ ...,\ {\nu}_{t,N_t}^a \} $, where ${\nu}_{t,i}^a = (t, b_t^i, \hat{f}_t^i)$ indicates the node of the $i$-th actor and $\hat{f}_t^i \in \mathbb{R}^C$ is the max-pooled feature of $f_t^i$. Similarly, we define the memory node set \emph{w.r.t.} ${v_t}$ as $V_t^m = \{ V_i^a \}_{i=t-L,i\neq t}^{t+L}$. Considering different levels of interactions, we link the actor nodes with three types of edges. More concretely, we define the first type as intra-actor edges $E_{intra}$, constructed by linking the nodes in the consecutive clips within the same tracklet. The remaining two edge sets are both inter-actor ones, which pass messages among different actors within the same clips and between different clips. Thus, we call them inter-actor intra-clip edges $E_{inter}^1$ and inter-actor inter-clip edges $E_{inter}^2$, respectively. $E_{inter}^1$ simply links all the node pairs within the same clips. To construct $E_{inter}^2$, with regard to an actor node, we add edges to all the memory nodes for message passing through actors in the long temporal range.

\subsubsection{\textbf{Single-layer Message Passing}}
Here, we show how to model interactions by message passing on the constructed graph with different types of edges. Compared to image-based tasks, interactions in action videos are more complex as we need to consider them in the spatio-temporal domain. For example, given two actors detected in different video clips, they can belong to either the same identity or different identities. Different identities can potentially enhance the final detection result based on their meaningful interactions, whereas actors belonging to the same identity cannot contribute to the interaction. However, as actors in one clip can represent the progress of an action at that specific timestamp, it is beneficial to model intra-actor relationships across different clips.

To be concrete, there are three ways to fulfill actor-actor interactions in one layer, namely intra-actor message passing, inter-actor intra-clip message passing and inter-actor inter-clip message passing, corresponding to $E_{intra}$, $E_{inter}^1$ and $E_{inter}^2$, respectively. Intra-actor message passing is based on the ordered temporal aggregator that captures dynamics across consecutive clips of the same identity. Both inter-actor message passing is attention-based, aiming at exploring interactions between different identities. 
Following previous works \cite{wu2019long}, we apply the self-attention block \cite{wang2018non} to model interactions. Fig. \ref{fig:3} (a) depicts the detailed structure, which shares key and value inputs. This design is adopted for all attention-based message passing. For simplicity, we only clarify the input of query and key in the following.

\noindent\textbf{Intra-actor Message Passing.} We propose the ordered temporal aggregator to model intra-actor message passing across different clips within the same tracklet. To achieve this, we merge actor node features in the adjacent clips via the edge set $E_{intra}$. Specifically, as shown in Fig. \ref{fig:3} (b), we use two fully-connected layers to fuse the node features from different directions rather than the position-independent self-attention. Then, we aggregate the two output features from the adjacent actor nodes.

\noindent\textbf{Inter-actor Intra-clip Message Passing.} The goal is to exploit actor-actor interactions within a video clip. As the graph is fully-connected at the timestamp, the adjacency matrix can be ignored without any negative influences. To be specific, the features of the actor node set $V_t^a$ can be stacked into one feature matrix $F_t \in \mathbb{R}^{1\times N_T \times C}$, which is then treated as both the query and key to compute the self-attention for node features.
In terms of the memory node set $V_t^m$, as the number of actors can be variant, we restrict the maximum number of actors to $N_{max}$ and generate the corresponding feature matrix $F_t^{M} \in \mathbb{R}^{2L \times N_{max} \times C}$. Then, we calculate the self-attention for the memory node features in a similar way.

\noindent\textbf{Inter-actor Inter-clip Message Passing.} This operation underlines actor-actor interactions between tracklets at different timestamps. We only update the features of the actor nodes $V_t^a$, as we aim to gradually enlarge the receptive field for memory nodes to establish the hierarchical learning structure. Since we connect all the memory nodes to actor nodes \emph{w.r.t.} $E_{inter}^2$, such message passing in graph nodes can be simply implemented by taking $F_t \in \mathbb{R}^{1\times N_T \times C}$ as the query and the reshaped $F_t^{M} \in \mathbb{R}^{1\times 2LN_{max} \times C}$ as the key.

\subsubsection{\textbf{Graph Down-sampling}}
\begin{figure*}[t]
   \centering
   \begin{subfigure}{0.3750\linewidth}
    \centering
    \includegraphics[width=\linewidth]{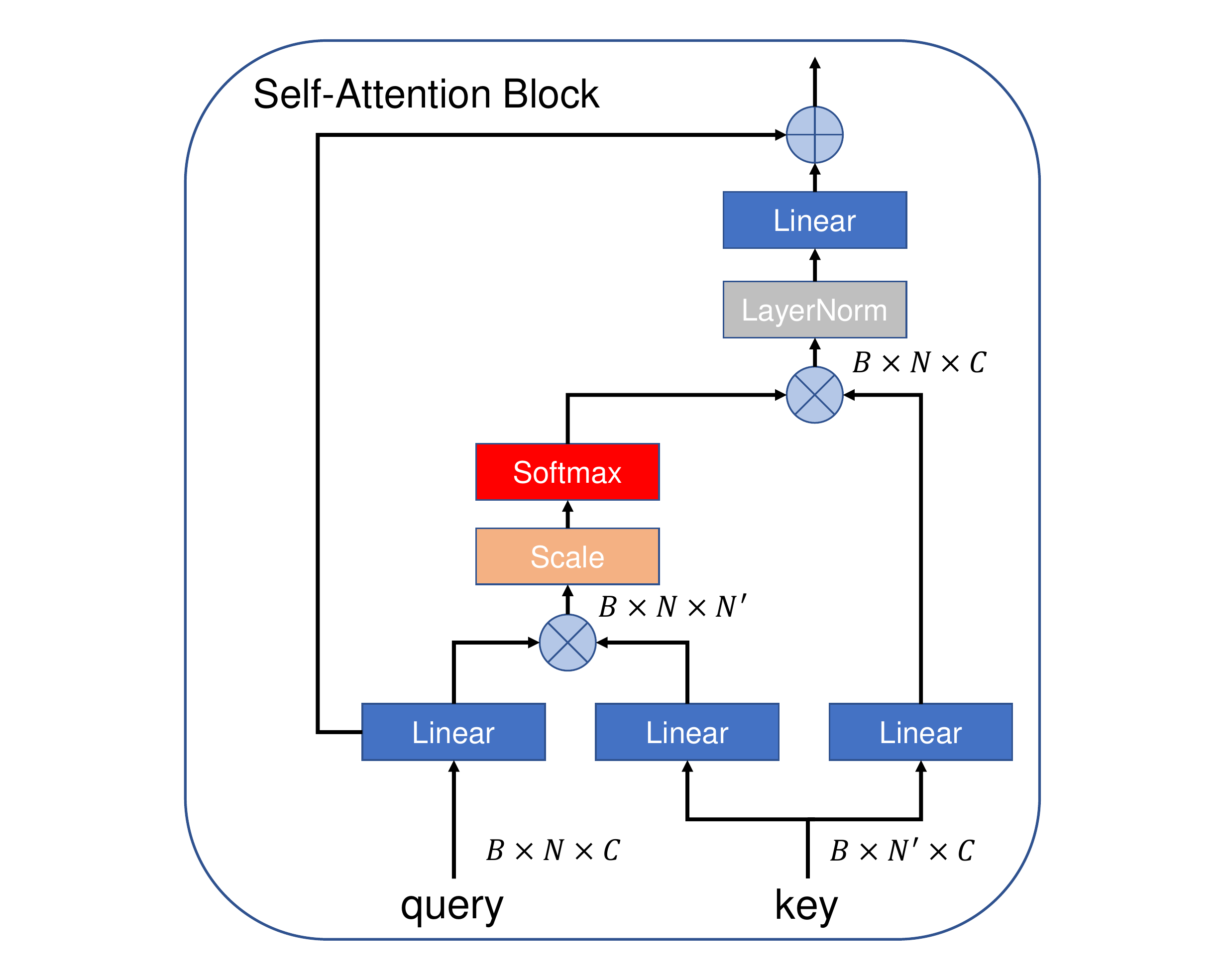}
    \caption{Self-attention block}
   \end{subfigure}%
   \begin{subfigure}{0.60\linewidth}
    \centering
    \includegraphics[width=\linewidth]{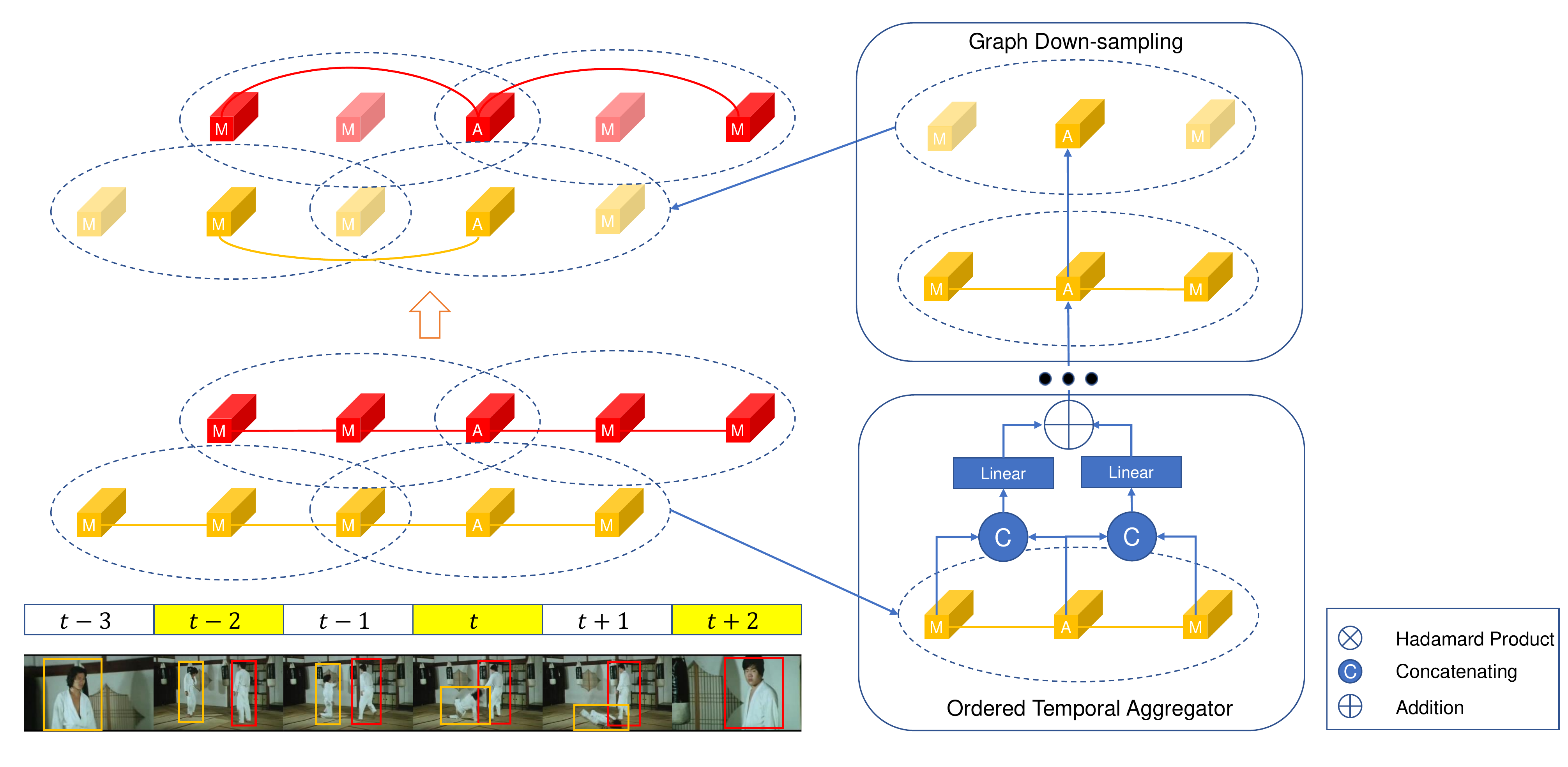}
    \caption{Identity-aware modeling}
   \end{subfigure}
   \vspace{-3mm}
  \caption{Message passing in HGNN. (a) Self-attention block used for graph message passing. (b) Identity-aware modeling. The nodes at timestamp $t$ belong to the actor node and the nodes at the timestamps in yellow color indicate the reference nodes. After down-sampling, transparent nodes (non-reference nodes) are removed from the graph.}
  \label{fig:3}
\end{figure*}

The ordered temporal aggregator only integrates features in temporally adjacent clips, thereby limiting capturing long-range relationships. 
To address this, we combine multi-scale context features in different layers via graph down-sampling.
As shown in Fig. \ref{fig:3} (b), after graph down-sampling, the linked nodes are merged. 
For any node in the graph, we first calculate its distance to the actor node through $E_{intra}$ (one node can access at most one actor node in this case) and take any accessible node with an even distance as the reference node. If two reference nodes link to the same non-reference one, we add a new edge that connects these two reference nodes to $E_{intra}$. Finally, the graph is down-sampled by removing all the non-reference nodes and their corresponding edges.

\begin{figure}[h]
    \centering
    \includegraphics[width=\linewidth]{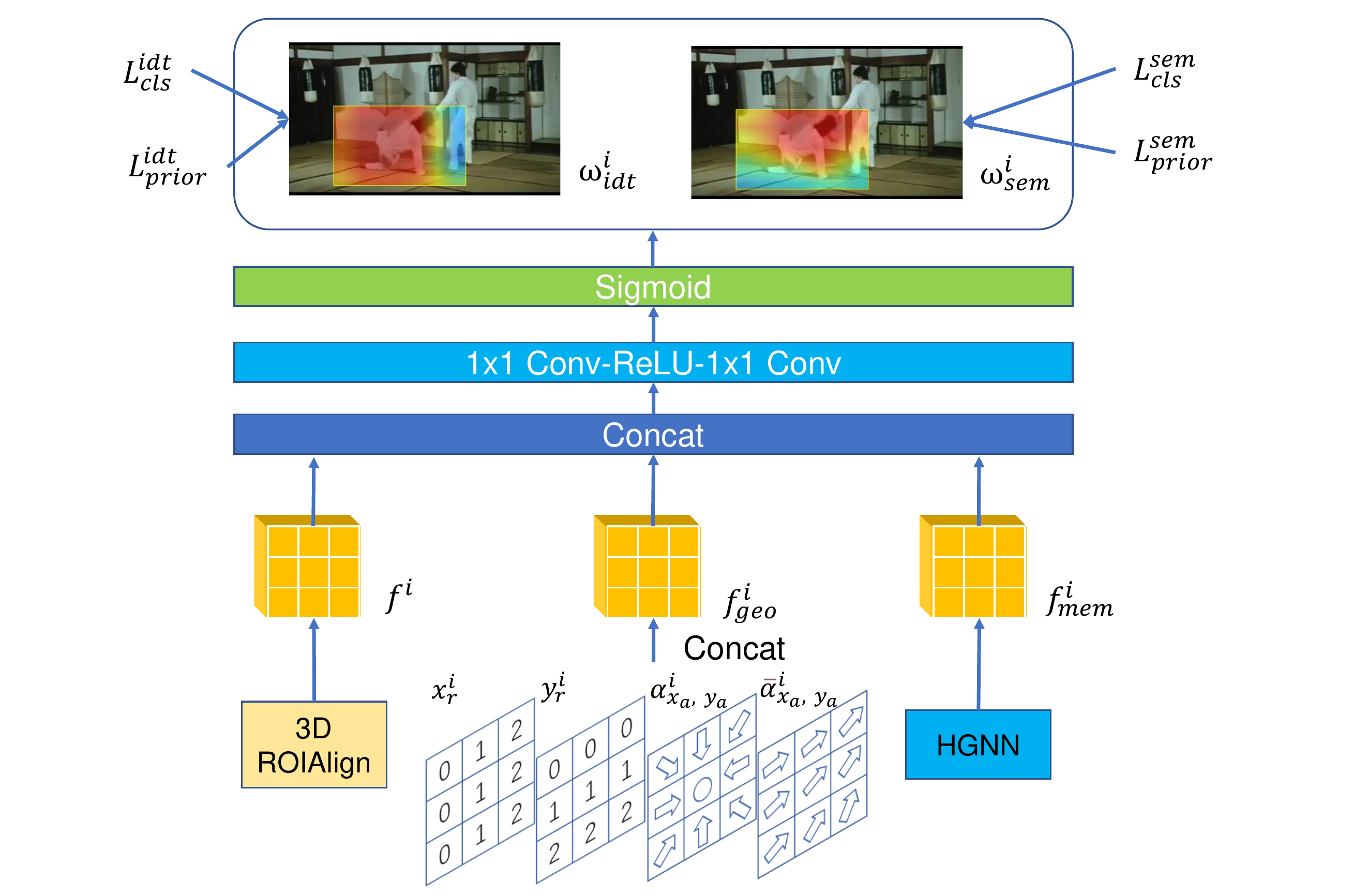}
    \vspace{-5mm}
  \caption{Dual attention module. We concatenate the actor features with geometry embedding as input. Several transforms are applied to generate dual attention maps supervised by the proposed classification and spatial prior losses.}
  \label{fig4}
\end{figure}

\subsection{Dual Attention Module}
Although HGNN models spatio-temporal relations, it is difficult to capture local details (based on the global-averaged feature maps), which are important in short-term context modeling \cite{li2019long, ulutan2020actor}. Besides, as a common practice in action detection, the detected bounding boxes are often expanded for more context \cite{wu2020context}. As a result, there usually exist multiple actors in one bounding box, some of whom may negatively affect the final action recognition result. To alleviate these weaknesses, we develop a novel dual attention module (DAM) to generate identity-related attention maps for action recognition. The key of DAM is disentangling one attention map into two sub-attention (\ie, semantic attention (SA) and identity attention (IA)) ones. In a bounding box, we define the central actor as the representative actor and any other actors as the interference ones. SA aims at locating regions with discriminative features for action recognition, no matter which actor it belongs to, while IA targets identity-related regions that belong to representative actors. The rationality behind using DAM is that the interference actors tend to influence the recognition of the representative actor's action; meanwhile, the standard classification loss imposed on the representative actor's feature probably inhibits the activation of the features belonging to interference actors, leading to inferior recognition results of their actions.
Fig. \ref{fig4} illustrates the framework of DAM, which is applied to the RoI-Aligned actor feature map $f^i \in \mathbb{R}^{C \times K \times K}$ without extra pooling operations\footnote{For simplicity, we neglect the subscript of timestamp $t$ in this subsection.}. Specifically, before the forward pass, we generate geometry embeddings as the spatial information. Taking the center coordinate of the $i$-th actor as the reference point, we represent the polar angle of the point $(x, y)$ as $\alpha_{x,y}^{i}$. The final geometry embedding of the relative spatial coordinate $(x_r^i, y_r^i)$ in the feature map $f_i$ is defined as: 

\begin{equation}\label{(1)}
   \begin{split}
      & f^i_{geo} = (x_r^i, y_r^i, \alpha_{x_a, y_a}^i, \bar{\alpha}_{x_a, y_a}^{i}) \\
      & x_a = x^i + x_r^i*h^i/K,\ y_a = y^i + y_r^i*w^i/K \\
      & \bar{\alpha}_{x_a, y_a}^{i} = \frac{1}{\sum_{j, j\neq i} p^j_{x_a, y_a} + \epsilon} 
      \sum_{j, j\neq i} p^j_{x_a, y_a} \alpha_{x_a, y_a}^j \\
   \end{split}
\end{equation}

\noindent where $(x^i, y^i, h^i, w^i)$ is the bounding box coordinate of the $i$-th actor, $(x_a, y_a)$ is the absolute coordinate corresponding to $(x_r^i, y_r^i)$, $\bar{\alpha}_{x_a, y_a}^{i}$ represents the polar angle of interference actors, $p^i_{x,y}$ is a binary indicator that equals 1 if $(x, y)$ is located in the bounding box of the $i$-th actor, and $\epsilon$ is used to avoid dividing by zero.

Then, the input to DAM is the concatenation of geometry embeddings, original features and memory features:

\begin{equation}\label{(2)}
    \begin{split}
        &f_{dam}^i = concate(f^i, f^i_{geo}, f^i_{mem})\\ 
        &\omega_{sem}^i = \sigma(FFN_{sem}(f_{dam}^i)),\
        \omega_{idt}^i = \sigma(FFN_{idt}(f_{dam}^i)) 
    \end{split}
\end{equation}

\noindent where $\sigma$ is the sigmoid function, $\odot$ indicates the Hadamard product, $\omega_{sem}^i$ and $\omega_{idt}^i$ denote the weight maps of SA and IA respectively, $f^i_{mem}$ refers to the memory features computed by simply averaging the output of HGNN in all the layers and applying one $1\times 1$ convolution layer for channel reduction, and $FFN_{sem}$ and $FFN_{idt}$ represent the attention functions, both of which are implemented by two continuous $1 \times 1$ convolution layers. Subsequently, we tile $f^i_{mem}$ in the spatial domain for fitness with concatenation. Finally, we calculate the attention-pooled features for classification as:

\begin{equation}\label{(3)}
    \vspace{-2.5mm}
   \begin{split}
      & f_{sem}^i = \frac{1}{K^2} \sum_{x=1}^K\sum_{y=1}^K f^i_{dam}(x,y) \odot \omega_{sem}^i(x,y)  \\
      & f_{idt}^{i} = \frac{1}{K^2} \sum_{x=1}^K\sum_{y=1}^K f^i_{dam}(x,y)\odot \bar{\omega}_{idt}^{i}(x,y)
   \end{split}
\end{equation}

\noindent where $(x,y)$ refers to the position of the weight map. Note that in practice, we adopt an improved version of $\omega_{idt}^i$, \ie, $\bar{\omega}_{idt}^{i} = \omega_{sem}^i \odot \omega_{idt}^i$, as the identity-related weight map for the representative actor, because using $\omega_{idt}^i$ alone tends to include none action-related regions. Moreover, to make the two attention modules focused, we add auxiliary losses considering the differences in classification and position prior in both the attentions. 

\noindent\textbf{Classification Loss.} As stated, SA locates regions with discriminative features for both representative and interference actors. To this end, we use a relaxed Binary Cross Entropy (BCE) loss as:

\begin{equation}\label{(4)}
   \begin{split}
      & L_{cls}^{sem} = \frac{1}{N}\sum_{i=1}^{N} \sum_{c=1}^{C}
      [y^i_c log \hat{y}^i_{c} + (1-E(y^i_c)) log(1-\hat{y}^i_{c})]  \\
      & E(y^i_c) = max(\{IOU(b^i, b^j) \cdot {y^j_c}\}_{j=1}^{N})
   \end{split}
\end{equation}

\noindent where $y^i$ is the ground truth, $\hat{y}^i$ is the prediction of $f_{sem}^i$, and $E$ represents the relaxation function. The relaxed BCE loss down-weights the negative sample in class $c$ as long as there exist interference actors belonging to the positive ones \emph{w.r.t.} this class. 
With $\omega_{idt}$, representative actor features are focused. We thus adopt the standard BCE loss for $f_{idt}^{i}$:

\begin{equation}\label{(5)}
   L_{cls}^{idt} = \frac{1}{N}\sum_{i=1}^{N} \sum_{c=1}^{C} 
      [y^i_c log \hat{y}^{i,idt}_c + (1-y^i_c) log(1-\hat{y}^{i,idt}_c)]
\end{equation}

\noindent where $\hat{y}^{i,idt}$ is the prediction of $f_{idt}^{i}$. The main difference between $L_{cls}^{sem}$ and $L_{cls}^{idt}$ lies in the weights of negative samples.

\noindent\textbf{Spatial Prior Loss.} 
As IA masks out the regions of interference actors, we add a spatial prior loss that makes the regions without interference actors highly activated: 

\begin{equation}\label{(6)}
   L_{prior}^{idt} = \frac{1}{K^2}\sum_{x=1}^K\sum_{y=1}^K (1-\hat{p}^i_{x,y}) (\omega_{idt}^i(x, y)-1)^2
\end{equation}

\noindent where $\hat{p}^i_{x,y}$ indicates whether there exist interference actors at the spatial position $(x, y)$ as:

\begin{equation}\label{(7)}
   \hat{p}^i_{x,y} = max(\{ p^j_{x,y} \}_{j\neq i})
\end{equation}

\noindent In practice, the bounding boxes are randomly scaled during training and we set $\hat{p}^i$ to 1 in the expanded area of $b^i$. 
As two attention maps are combined by multiplication, the weights of SA influence the learning of IA. If the weights of interference actors are too low, IA does not need to down-weight these regions. To better guide the training of both attentions, we also add a spatial prior loss on SA based on the ranking loss with a small margin $\gamma=0.1$ to make the regions with interference actors more focused:

\begin{equation}\label{(8)}
   \begin{split}
      L_{prior}^{sem} =\ max(0,\ 
      &\gamma - \frac{1}{K^2}(\sum_{x=1}^K\sum_{y=1}^K \hat{p}^i_{x,y} \cdot \omega_{sem}^i(x,y) \\
      &-\sum_{x=1}^K\sum_{y=1}^K (1-\hat{p}^i_{x,y}) \cdot \omega_{sem}^i(x,y))) \\
   \end{split}
\end{equation}

The spatial prior losses imposed on both attentions indeed facilitates DAM in locating identity-related regions, \ie, SA has higher weights on the regions with interference actors and IA has complete freedom to decide which identity the current region belongs to.


\subsection{Feature Fusion}
As aforementioned, HGNN captures long-term inter-actor and intra-actor interactions and DAM exploits local details in short-term context. Finally, we fuse the outputs from the two branches to build final representation. The fused feature maps are further fed to two convolutional layers to generate the action score. We apply the standard BCE Loss as in Eq. \eqref{(5)} to compute the training loss $L_{cls}^{fuse}$:

\begin{equation}\label{(9)}
 L_{cls}^{fuse} = \frac{1}{N}\sum_{i=1}^{N} \sum_{c=1}^{C} 
      [y^i_c log \hat{y}^{i,fuse}_{c} + (1-y^i_c) log(1-\hat{y}^{i,fuse}_{c})]
\end{equation}

\noindent where $\hat{y}^{i,fuse}_{c}$ indicates the prediction of the fused feature $f_{fuse}^i = f_{mem}^i + f_{idt}^{i}$. Here, $f_{mem}^i$ is the output of HGNN.
\subsection{Loss Function} 
The final loss is the weighted sum of the losses above:

\begin{equation}\label{(10)}
   L= L_{cls}^{fuse} + \lambda_{aux}(L_{cls}^{sem} + L_{cls}^{idt} + L_{prior}^{sem} + L_{prior}^{idt})
\end{equation}

\noindent where $\lambda_{aux}$ is a trade-off hyper-parameter between different losses.

\newsavebox\tableb
\newsavebox\tablee 
\savebox{\tableb}{
   \begin{tabular}{l|c}
      model & mAP \\
      \hline\hline
      Baseline & 26.52 \\
      \hline
      +HGNN & 30.76 \\
      +HGNN(frozen) & 30.19 \\
   \end{tabular}%
}
\savebox{\tablee}{
   \begin{tabular}{l|c}c
      model & mAP \\
      \hline\hline
      HGNN & 30.76 \\
      \hline
      +Dual(1x) & ? \\
      +Dual(1.3x) & ? \\
      +Dual(1.5x) & ?
   \end{tabular}%
}

\begin{table*}[thbp]
   \small
   \centering
   \caption{Experimental results (\%) in terms of different ablation studies.}
   \subcaptionbox{Hierarchical intra-actor modeling.}[0.3\linewidth]{
      \centering
      \vbox to \ht\tableb{%
         \begin{tabular}{c|cc}
            Stages & HGNN & HGNN-s \\
            \hline\hline
            1 & 29.32 & 28.92 \\
            2 & 30.11 & 29.26 \\
            3 & 30.76 & 29.21 \\
           
         \end{tabular}

      }
                \label{tab:1a}
   }
   \subcaptionbox{Graph message passing.}[0.3\linewidth]{
      \centering
      \vbox to \ht\tableb{%
         \begin{tabular}{l|c}
            Model & mAP \\
            \hline\hline
            Baseline & 26.52 \\
            \hline
            +HGNN & 30.76 \\
            +HGNN (frozen) & 30.19 \\
         \end{tabular}%
      }
      \label{tab:1b}
   }
   \subcaptionbox{Graph learning.}[0.3\linewidth]{
      \centering
      \vbox to \ht\tableb{%
         \begin{tabular}{l|c}
            Model & mAP \\
            \hline\hline
            Baseline & 26.52 \\
            \hline
            +HGNN (GCN) & 28.95 \\
            +HGNN (ours) & 30.76 \\
         \end{tabular}%
      }
   }
   \vfill
   \subcaptionbox{Dual attention.}[0.3\linewidth]{
      \centering
      \vbox to \ht\tablee{%
         \begin{tabular}{l|c}
            Model & mAP \\
            \hline\hline
            Baseline+HGNN & 30.76 \\
            \hline
            +Semantic & 30.57 \\
            +Identity & 30.35 \\
            IGMN & 31.22 \\
         \end{tabular}%
      }
   }
   \subcaptionbox{Auxiliary losses.}[0.3\linewidth]{
      \centering
      \vbox to \ht\tablee{%
         \begin{tabular}{l|c}
            Model & mAP \\
            \hline\hline
            IGMN & 31.22 \\
            \hline
            IGMN (w/o cls) & 30.51 \\
            IGMN (w/o prior) & 30.91 \\
            IGMN (w/o aux) & 30.33 \\
         \end{tabular}%
      }
   }
  \subcaptionbox{Local context.}[0.3\linewidth]{
      \centering
      \vbox to \ht\tablee{%
         \begin{tabular}{l|c}
            Model & mAP \\
            \hline\hline
            Baseline+HGNN & 30.76 \\
            \hline
            IGMN (1$\times$) & 31.10 \\
            IGMN (1.25$\times$) & 31.22 \\
            IGMN (1.5$\times$) & 31.15
         \end{tabular}%
      }
  }
   \label{tab:temps}
\end{table*}

\begin{figure*}[htbp]
  \centering
  \setlength\tabcolsep{1pt}%
  \begin{tabular}[c]{rcccccc}
    Original &
    \multicolumn{1}{l}{
      \begin{subfigure}{0.15\linewidth}
        \includegraphics[height=1.5cm, width=\linewidth]{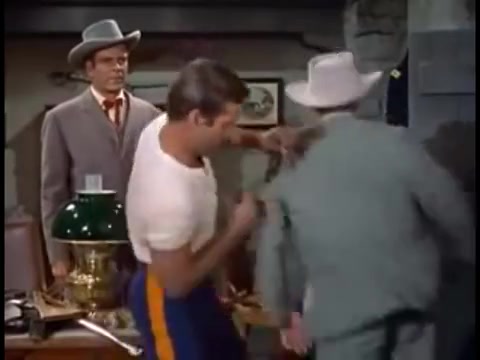}
     \end{subfigure}
    } &
    
    \multicolumn{1}{l}{
      \begin{subfigure}{0.15\linewidth}
        \includegraphics[height=1.5cm, width=\linewidth]{figures/_Z6_ori.jpg}
     \end{subfigure}
    } &
    
    \multicolumn{1}{l}{
      \begin{subfigure}{0.15\linewidth}
        \includegraphics[height=1.5cm, width=\linewidth]{figures/_Z6_ori.jpg}
     \end{subfigure}
    } &

    \multicolumn{1}{l}{
      \begin{subfigure}{0.15\linewidth}
        \includegraphics[height=1.5cm, width=\linewidth]{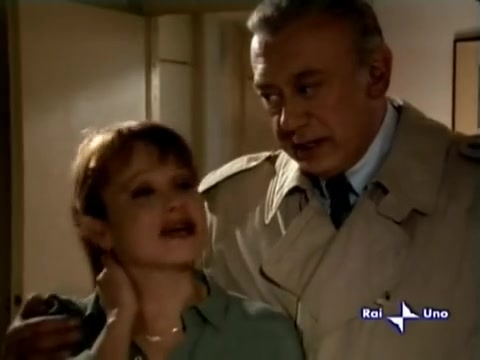}
     \end{subfigure}
    } &
    
    \multicolumn{1}{l}{
      \begin{subfigure}{0.15\linewidth}
        \includegraphics[height=1.5cm, width=\linewidth]{figures/oq_ori.jpg}
     \end{subfigure}
    } &
    
    \multicolumn{1}{l}{
      \begin{subfigure}{0.15\linewidth}
        \includegraphics[height=1.5cm, width=\linewidth]{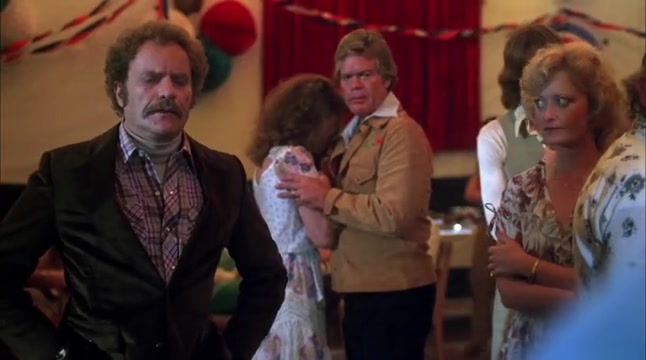}
     \end{subfigure}
    } \\
    \specialrule{0em}{1pt}{1pt}
    
    Semantic &
    \begin{subfigure}{0.15\linewidth}
      \includegraphics[height=1.5cm, width=\linewidth]{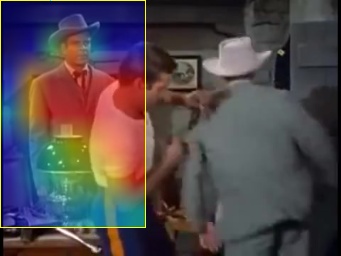}
    \end{subfigure}&

    \begin{subfigure}{0.15\linewidth}
      \includegraphics[height=1.5cm, width=\linewidth]{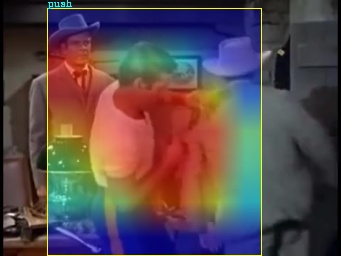}
    \end{subfigure}&

    \begin{subfigure}{0.15\linewidth}
      \includegraphics[height=1.5cm, width=\linewidth]{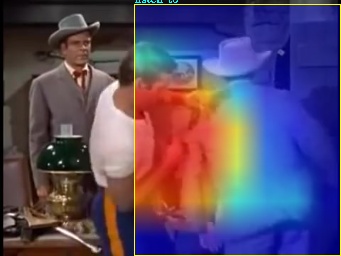}
    \end{subfigure} &

    \begin{subfigure}{0.15\linewidth}
      \includegraphics[height=1.5cm, width=\linewidth]{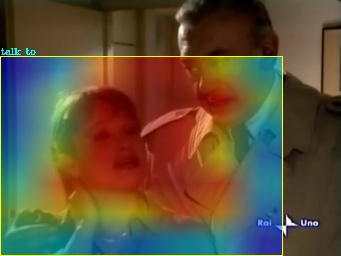}
    \end{subfigure}&

    \begin{subfigure}{0.15\linewidth}
      \includegraphics[height=1.5cm, width=\linewidth]{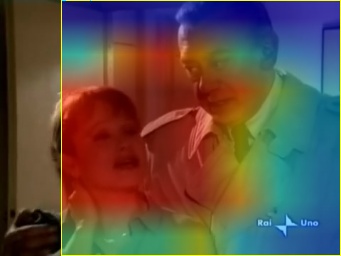}
    \end{subfigure}&

    \begin{subfigure}{0.15\linewidth}
      \includegraphics[height=1.5cm, width=\linewidth]{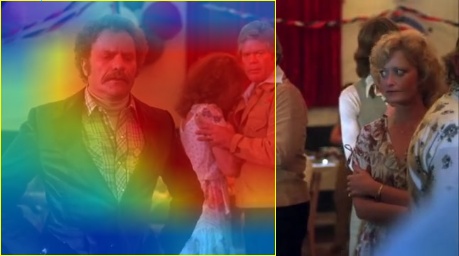}
    \end{subfigure} \\

    \specialrule{0em}{1pt}{1pt}

    Identity &
    \begin{subfigure}{0.15\linewidth}
      \includegraphics[height=1.5cm, width=\linewidth]{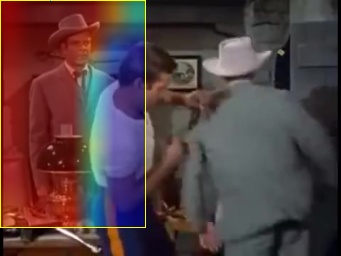}
    \end{subfigure}&

    \begin{subfigure}{0.15\linewidth}
      \includegraphics[height=1.5cm, width=\linewidth]{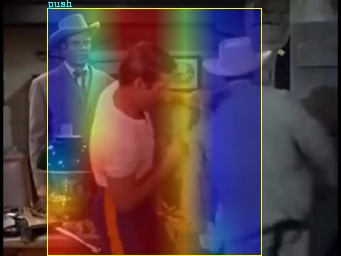}
    \end{subfigure}&

    \begin{subfigure}{0.15\linewidth}
      \includegraphics[height=1.5cm, width=\linewidth]{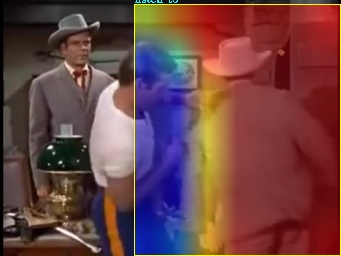}
    \end{subfigure} &

    \begin{subfigure}{0.15\linewidth}
      \includegraphics[height=1.5cm, width=\linewidth]{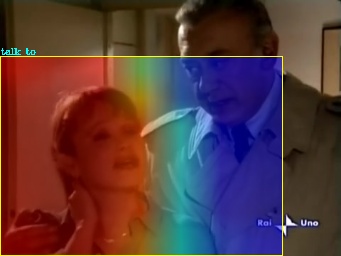}
    \end{subfigure}&

    \begin{subfigure}{0.15\linewidth}
      \includegraphics[height=1.5cm, width=\linewidth]{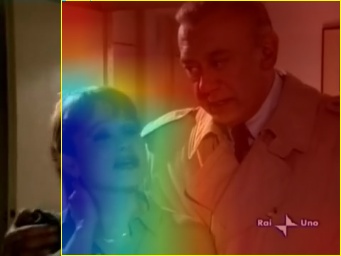}
    \end{subfigure}&

    \begin{subfigure}{0.15\linewidth}
      \includegraphics[height=1.5cm, width=\linewidth]{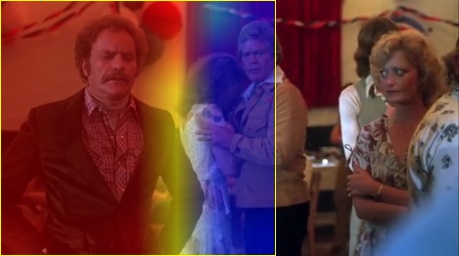}
    \end{subfigure} \\

    \specialrule{0em}{0pt}{0pt} 
    Label & stand/watch & stand/push & stand
    & stand/talk & stand/watch/listen & stand/watch \\
  
  \end{tabular}  
  \caption{Visualization of DAM. Each column shows the attention maps of an actor and the corresponding action labels.}
  \label{fig:5}
\end{figure*}

\section{Experimental Results}
\subsection{Evaluation Setup}
\noindent\textbf{Dataset.} 
Evaluation is made on the challenging AVA dataset \cite{gu2018ava}, which requires context to recognize actions. It contains 235 training and 64 validation videos. Instance-level annotations with bounding boxes and 80 atomic action classes are provided on sparsely sampled key frames at 1 FPS. 
There are three kinds of actions, \ie, pose actions, person-person interactions, and person-object interactions. We employ both v2.1 and v2.2 in comparison to the state-of-the-arts and v2.2 for ablation study. Following the official protocol, we adopt the frame-level mean Average Precision (mAP) at the IoU threshold of 0.5 as the evaluation metric. The detection results are reported based on the top-60 most common actions.

\noindent\textbf{Person Detector and Tracker.}
We adopt the same actor proposals as in \cite{tang2020asynchronous} for fair comparison. Spatial bounding boxes are generated on the key frame of each clip by Faster R-CNN \cite{ren2015faster} with ResNeXt-101-FPN as the backbone. The model is pre-trained on ImageNet \cite{deng2009imagenet} and MS-COCO \cite{lin2014microsoft} and then fine-tuned on AVA. We use the tracker in \cite{zhang2019structured} to link detected actors.

\noindent\textbf{Video Backbone.}
We choose SlowFast \cite{feichtenhofer2019slowfast} as the backbone. Following previous works, we double the spatial resolution of $res_5$ by replacing the down-sampling layer with dilated convolutions. In ablation study, we adopt SlowFast-50 instantiated with the sampled input 4$\times$16 pre-trained on the Kinetics-700 dataset.

\noindent\textbf{Training and Inference.}
In training, we set $\lambda_{aux}$ to 0.5 and $\gamma$ to 1. Since a person only belongs to one pose action, we apply the softmax function instead of sigmoid to generate scores for pose action classes and ignore the label when calculating the auxiliary loss \emph{w.r.t.} DAM. As for updating the memory bank, we use the asynchronous memory update algorithm \cite{tang2020asynchronous} that directly saves the forward actor features in the memory bank. The input clips are rescaled such that the shortest side is 256, and different clips are stacked to form the batch by zero paddings to the maximum size. We only use ground-truth bounding boxes for training and randomly jitter actor proposals for data augmentation. The proposals are further scaled to at most 1.5$\times$ for sufficient training. $L$ is set to 30 for memory features. We use the stochastic gradient descent (SGD) optimizer with batch size 12, initial learning rate 0.003, weight decay 0.0005, and momentum 0.9. The training process runs for 140k iterations, and the learning rate is decreased by a factor of 10 at iterations 93k and 120k. In testing, the detected human boxes with the confidence score larger than 0.8 are selected, and we enlarge their scales by 1.25 for more context. The shorter side of the input frames is scaled to 256, identical to the setting in training.

\subsection{Ablation Study}

\noindent\textbf{Hierarchical Intra-actor Modeling.}
We discard DAM and only use the features from HGNN for recognition.
We evaluate intra-actor modeling in the hierarchical structure by adopting different numbers of graph layers. For more comprehensive analysis, we add a simpler version of HGNN, \ie, HGNN-s, in which ordered temporal aggregator and graph down-sampling are removed.
As in Table \ref{tab:temps} (a), more layers usually lead to better results for both HGNN and HGNN-s, indicating the credit of multi-level graph learning. Besides, HGNN benefits more from deeper layers due to enlarged receptive fields between consecutive layers.
In addition, the model performs better when intra-actor modeling is applied.
Such results demonstrate its effectiveness.

\noindent\textbf{Graph Message Passing.}
Unlike previous methods \cite{wu2019long} that freeze the features in the memory network, our model passes messages through the memory graph with intra-clip self-attention and inter-clip self-attention. To validate the message passing in HGNN, we implement a variant by removing the feature updating procedure for memory nodes. Table \ref{tab:temps} (b) shows the comparative results, indicating that graph message passing leads to improvement over the counterpart with frozen memory nodes.

\begin{figure*}[!ht]
    \centering
    \includegraphics[width=\linewidth]{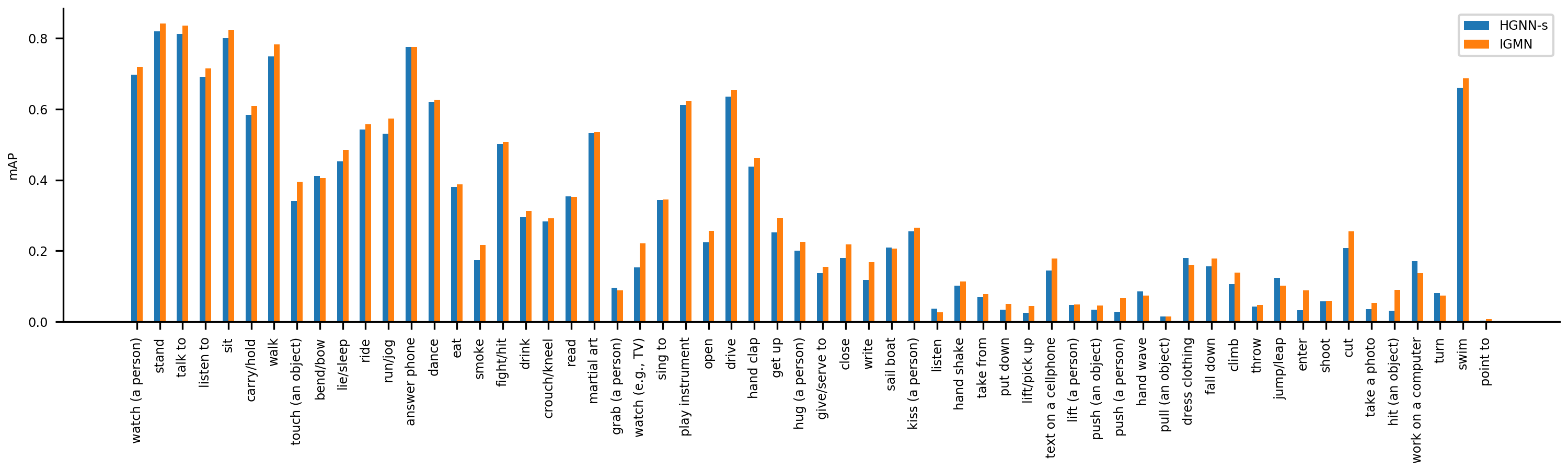}
    \vspace{-8mm}
  \caption{Impact of identity-aware modeling by comparison
  between HGNN-s
  and IGMN in terms of per-class APs on AVA. 
  }
  \vspace{-1mm}
\label{fig:6}
\end{figure*}

\begin{table*}[htbp]
  \caption{Comparison with state-of-the-art methods.
  Note that we do not include the results with multiple scales and flips.
  }
  \vspace{-3mm}
  \begin{tabular}{l|c|c|c|c|c|c|c|c}
    Method & Reference & Memory & Backbone & Pretrain & Input & Modality & AVA & Val. mAP \\
    \hline\hline
    RelationGraph \cite{zhang2019structured} & CVPR'19 & $\times$ & R50-NL & K400 & 36 x 1 & V & v2.1 & 22.2\\
    SlowFast \cite{feichtenhofer2019slowfast} & ICCV'19 & $\times$ & SlowFast-R50,4x16 
    & K400 & 64 x 1 & V & v2.1 & 24.2 \\
    LFB \cite{wu2019long} & CVPR'19 & \checkmark & R50-NL 
    & K400 & 32 x 2 & V & v2.1 & 25.8 \\
    LSTR \cite{li2019long} & MM'19 & \checkmark & R101 & ImageNet & 16 x 1 & V + F & v2.1 & 27.2\\  
    Context-Aware RCNN \cite{wu2020context} & ECCV'20 & \checkmark & R50-NL
    & K400 & 32 x 2 & V & v2.1 & 28.0 \\
    ACAR-Net \cite{pan2020actor} & CVPR'21 & \checkmark & SlowFast-R50,8x8
    & K400 & 32 x 2 & V & v2.1 & 28.3 \\
    AIA \cite{tang2020asynchronous} & ECCV'20 & \checkmark & SlowFast-R50,4x16 
    & K700 & 32 x 2 & V & v2.1 & 28.9 \\
    \hline
    \textbf{IGMN} & Ours & \checkmark & SlowFast-R50,4x16 & K700 & 32 x 2 & V & v2.1 & \textbf{30.2} \\
    \hline \hline
    SlowFast \cite{feichtenhofer2019slowfast} & ICCV'19 & $\times$ & SlowFast-R101,8x8 
    & K700 & 32 x 2 & V & v2.2 & 29.3 \\
    AIA \cite{tang2020asynchronous} & ECCV'20 & \checkmark & SlowFast-R50,4x16 
    & K700 & 32 x 2 & V & v2.2 & 29.8 \\
    AIA \cite{tang2020asynchronous} & ECCV'20 & \checkmark & SlowFast-R101,8x8 
    & K700 & 32 x 2 & V & v2.2 & 32.2 \\
    \hline
    \textbf{IGMN} & Ours & \checkmark & SlowFast-R50,4x16 & K700 & 32 x 2 & V & v2.2 & 31.2 \\
    \textbf{IGMN} & Ours & \checkmark & SlowFast-R101,8x8 & K700 & 32 x 2 & V & v2.2 & \textbf{33.0} \\
  \end{tabular}
  \label{tab:2}
\end{table*}

\noindent\textbf{Graph Learning.}
A common way in current methods \cite{zhang2019structured} is to aggregate node features within a tracklet based on graph convolution network (GCN) \cite{kipf2016semi}. To highlight the advantage of intra-actor modeling, we adopt graph convolutions for intra-actor modeling. The results are displayed in Table \ref{tab:temps} (c), where graph convolutions perform worse than ordered temporal aggregation and graph down-sampling. It is because all the nodes within the same tracklet are connected with each other when using GCN. 

\noindent\textbf{Dual Attention.}
In DAM, SA seeks for discriminative regions in the feature maps, while IA tries to generate identity-aware masks. As in Table \ref{tab:temps} (d), better results are reached by making use of both attentions than either of them, proving the effectiveness. It is also visualized in the qualitative results \emph{w.r.t.} the attention maps by DAM in Fig. \ref{fig:5}, where discriminative regions are located in bounding boxes by SA and identity-irrelevant ones are masked out by IA. Taking the second column as an example, the head of the first person and the body of the second person are captured by SA, corresponding to ``watch'' and ``push'', respectively. Moreover, IA down-weights the activated regions of the interference actors to avoid disturbance. If only SA is applied, the target of the second person (without ``watch'') inhibits the activation of the head region of the first person; thus, there are likely to be some conflicts during optimization. This observation confirms the importance of DAM.


\noindent\textbf{Auxiliary Losses.}
We analyze the effect of the auxiliary losses in DAM, which contain a classification loss and a spatial prior loss designed for disentangling the semantic and identity cues. As in Table \ref{tab:temps} (e), both the auxiliary losses improve the results, where the classification loss contributes more. This also shows the problem of incorrect label assignment for samples with interference actors where separate classification target indeed takes effect.

\noindent\textbf{Local Context.}
Since HGNN is applied to pooled feature maps, identity-aware local context information is not well explored. Similar to \cite{wu2020context}, we enlarge the context by adjusting the box size. The results are displayed in Table \ref{tab:temps} (f), from which we can observe that larger box scales lead to slightly better scores.

\noindent\textbf{Identity-aware Modeling.}
Both HGNN and DAM are designed for identity-aware modeling. To illustrate the necessity of identity constraint, we show category-wise results and compare them with those of HGNN-s, by removing intra-actor modeling and DAM. As in Fig. \ref{fig:6}, identity modeling benefits most action categories (46/60), especially for the ones requiring identity clues across clips (\eg, ``open'', ``falling down'') or local details (\eg, ``push'', ``hug'').

\subsection{Comparison to State-of-the-Art Methods}
We compare our method with the state-of-the-art ones on the validation set of AVA in Table \ref{tab:2}. For comparison with early works, we also use v2.1 to train IGMN on the SlowFast R50 backbone. Under the similar setting, \ie, using the same SlowFast backbone pre-trained on Kinetics-700 and the same actor proposals, our method outperforms AIA \cite{tang2020asynchronous} with different backbones. Notably, we achieve 33.0\% in mAP by single-scale testing based on SlowFast-R101, establishing the new state of the art on AVA. In addition, an extra object detection step is used in AIA to find human-object interactions, which may bring further gains to our framework as well.

\section{Conclusion}
This paper proposes IGMN for action detection, with HGNN exploiting actor relations for long-term context and DAM modeling identity-aware discriminative features for short-term context. Extensive experiments show that IGMN achieves a significant performance gain on the challenging AVA dataset. Moreover, our method reveals the importance of incorporating identity information into action detection, providing valuable insights into future work.

\section*{ACKNOWLEDGMENTS}
This work is partly supported by the National Natural Science Foundation of China (No. 62022011), the Research Program of State Key Laboratory of Software Development Environment (SKLSDE-2021ZX-04), and the Fundamental Research Funds for the Central Universities.

\bibliographystyle{ACM-Reference-Format}
\balance
\bibliography{egbib}

\end{document}